\documentclass[10pt,twocolumn,letterpaper]{article}

\usepackage{iccv}
\usepackage{times}
\usepackage{epsfig}
\usepackage{graphicx}
\usepackage{amsmath}
\usepackage{amssymb}
\usepackage{bbm}

\usepackage{subfiles}
\usepackage{comment}
\usepackage{multirow}
\usepackage{makecell}
\usepackage{adjustbox}
\usepackage{tabularx}
\usepackage{soul}
\usepackage[title]{appendix}
\usepackage{placeins}
\DeclareMathOperator*{\argmax}{arg\,max}

\usepackage[pagebackref=true,breaklinks=true,letterpaper=true,colorlinks,bookmarks=false]{hyperref}

\iccvfinalcopy 

\def\iccvPaperID{*} 

\ificcvfinal\fi

\begin{document}

\title{
Harnessing Unrecognizable Faces for Improving Face Recognition 
}
\author{Siqi Deng \quad 
Yuanjun Xiong \quad 
Meng Wang \quad \
Wei Xia\quad \
Stefano Soatto \\
Amazon AWS AI\\
{\tt\small \{siqideng, yuanjx, mengw, wxia, soattos\}@amazon.com}
}

\maketitle

\ificcvfinal\thispagestyle{empty}\fi

\begin{abstract} 
The common implementation of face recognition systems as a cascade of a detection stage and a recognition or verification stage can cause problems beyond failures of the detector. When the detector succeeds, it can detect faces that cannot be recognized, no matter how capable the recognition system. Recognizability, a latent variable, should therefore be factored into the design and implementation of face recognition systems. We propose a measure of recognizability of a face image that leverages a key empirical observation: An embedding of face images, implemented by a deep neural network trained using mostly recognizable identities,  induces a partition of the hypersphere whereby unrecognizable identities {\em cluster together}. This occurs regardless of the phenomenon that causes a face to be unrecognizable, be it optical or motion blur, partial occlusion, spatial quantization, poor illumination. Therefore, we use the distance from such an ``unrecognizable identity'' as a measure of recognizability, and incorporate it in the design of the overall system. We show that accounting for recognizability reduces the error rate of single-image face recognition by 58\% at FAR=1e-5 on the IJB-C Covariate Verification benchmark, and reduces the verification error rate by 24\% at FAR=1e-5 in set-based recognition on the IJB-C benchmark.
\end{abstract}

\section{Introduction}
\label{sec:introduction}

\begin{figure}[!t]
\begin{center}
\includegraphics[width=\linewidth]{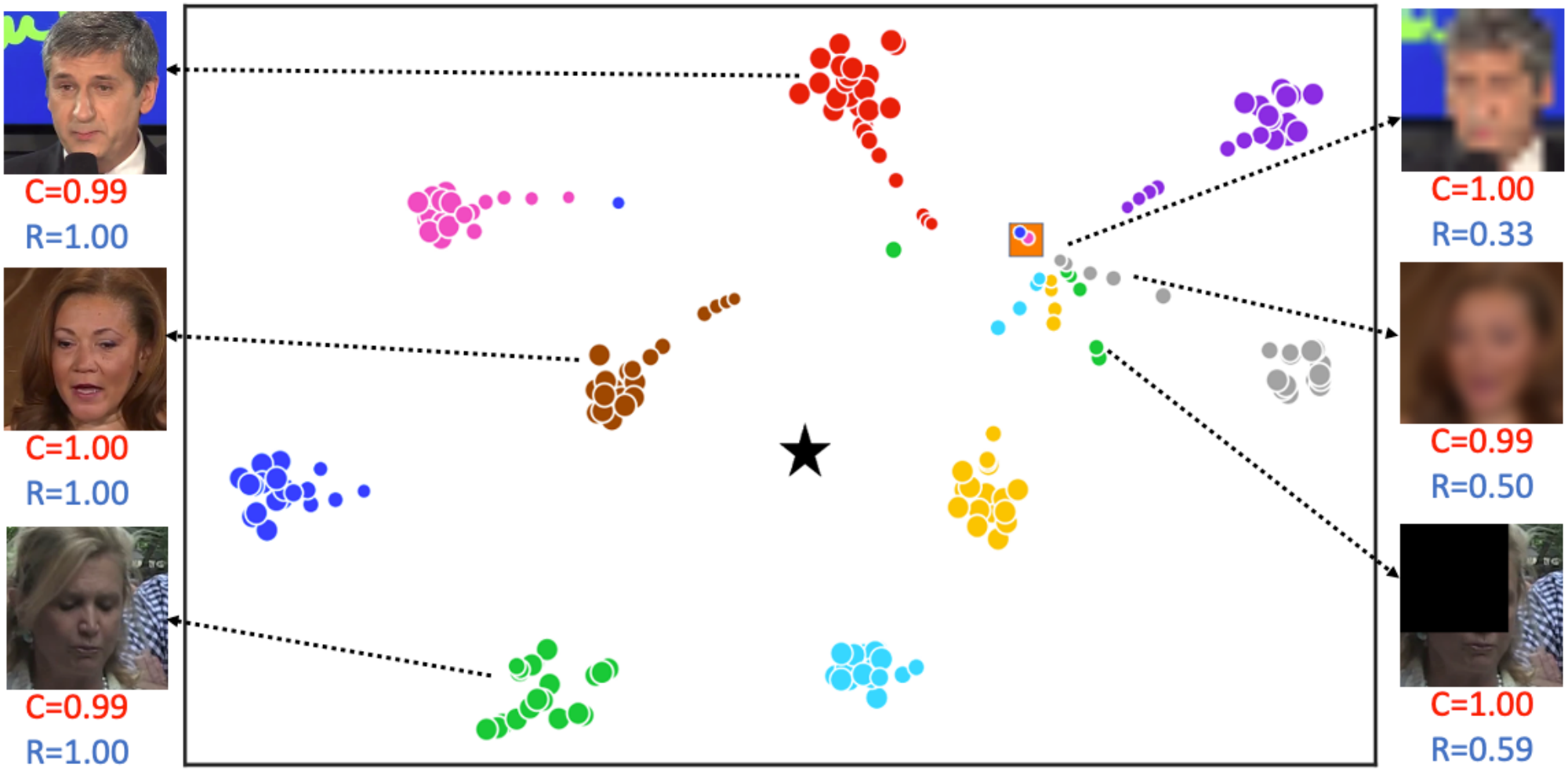}
\end{center}
  \caption{\label{fig_tsne_trajectory}
Hypersphere embeddings~\cite{wang2018cosface} of different faces from the IJB-C dataset (colored clusters) visualized as circles with area proportional to recognizability, using t-SNE~\cite{maaten2008visualizing}. As the images are perturbed artificially, becoming increasingly unrecognizable, their embeddings migrate to join a common cluster (orange square). Such an ``unrecognizable identity'' (UI) is described in Sec.~\ref{RS} and distinct from the centroid of the recognizable embeddings (black pentagram). Note the difference between {\em face detection confidence (C)} and {\em embedding based recognizability score (R)}. The former is the output of a face detector and measures the likelihood that the image contains a face while the latter measures if the face can be recognized.
}
\end{figure}

We aim at making face recognition systems easier to use responsibly. This requires not just reducing the error rate, but also producing interpretable performance metrics, with estimates of when recognition can be performed reliably, or otherwise should not be attempted, or deemed unreliable. 

In most face recognition systems, each image is first fed to a face detector (FD), that returns the location and shape of a number of bounding boxes likely to portray faces. Those, together with the image, are then fed to a downstream face recognition (FR) module that returns either one of $K$ labels corresponding to identities in a database (search), or a binary label corresponding to whether or not the bounding box matches a given identity  (verification).  FD and FR  are typically non-interacting modules trained on different datasets: The FD is tasked with finding faces no matter whether they are recognizable. The FR is tasked with mapping each detection onto one of $K$ identities. An obvious failure mode of such cascaded systems is when the FD is {\em wrong}: If a face is not detected, obviously it cannot be correctly recognized. If a detected bounding box does not show a face, the FR system will nonetheless map it to one of the known identities, unless post-processing steps are in place, typically involving a threshold on some confidence measure.

But even when the FD is {\em right}, there remain the following problems:

First, while an image may contain enough information to decide that there {\em is} a face, it may not contain enough to determine {\em whose} face it is, regardless of how good the FR system is. This creates a gap between the FD, tasked to determine that there is a face regardless of whether it is recognizable, and the FR, tasked to recognize it. An FR system should not try to recognize a face that is not recognizable.
\footnote{
An optimal (Bayesian) FR system would forgo the FD and marginalize over all possible locations and shapes, which is obviously intractable. But conditioning on the presence of a face, rather than marginalizing it, is not the only problem: There is another latent variable, {\em ``recognizability''} that is unaccounted for and instead assumed to be true by the FR. 
} Accordingly, {\em how can we measure and account for the recognizability of a face image in face recognition?}

Second, failure to take into consideration recognizability can lead to misleading results in face recognition benchmarks. Unrecognizable faces due to optical, atmospheric, or motion blur, spatial quantization, poor illumination, partial occlusion, etc, are typically used to train and score FD systems, but FR systems are trained using recognizable faces, lest one could not establish ground truth. Accordingly, {\em how can we  balance the reward in detecting unrecognizable faces with the risk in failure to recognize them?}
    
Third, failure to account for the recognizability of a face can have consequences beyond the outcome of FR on that face. Consider the problem of set-based face recognition where the identity is to be assigned {\em not} to a single image, but to a small collection of images known to come from the same identity,  some of which unrecognizable. These may be frames of a video, some of which affected by motion blur or partial occlusions. 
It may seem that using all the available data can only improve the quality of the decision. However, uniform averaging can significantly degrade performance.  Accordingly, {\em how should one combine images in set-based face recognition, assuming we have available a measure of recognizability?}

\begin{figure}[!t]
\begin{center}
\includegraphics[width=\linewidth]{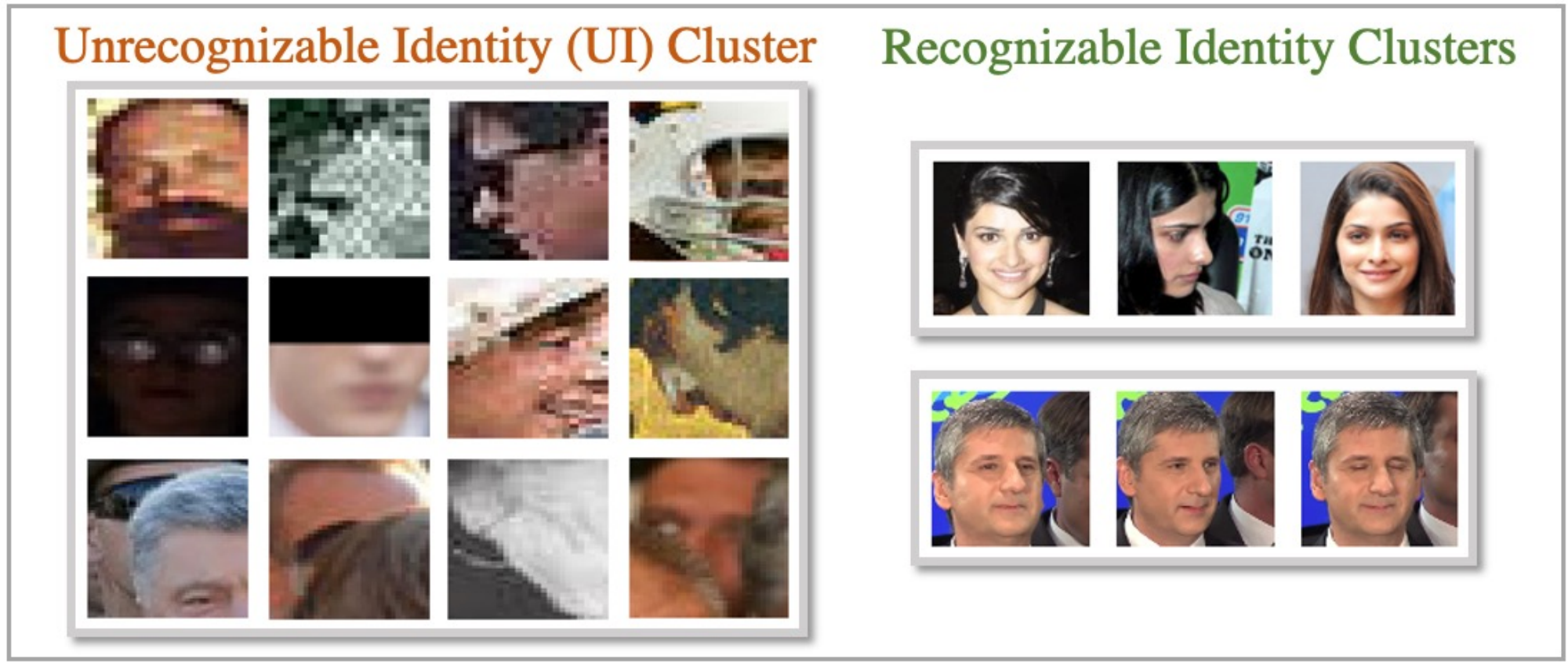}
\end{center}
  \caption{\label{face_clusers} During face clustering, faces with low recognizability scores are grouped into one cluster by distance based clustering method (left), compared with clusters of faces with known identities (right). We refer to this cluster as unrecognizable identity cluster (UI) cluster. The UI cluster including images subject to occlusion, optical or motion blur, low resolution, poor illumination, etc. So the clustering of UIs is not simply due to visual similarity. }
\end{figure}

\subsection{Main Hypothesis and Empirical Observation}

The three questions above point to the need to explicitly represent ``recognizability'' as a latent variable. As was the case for the FD,  marginalization is impractical. We instead hypothesize that recognizability can be quantified inferentially. A measure of recognizability could then be used to complete the hypothesis space for FR, effectively adding an additional class for {\em unrecognizable identities}, akin to open-set classification. An estimate of recognizability would also allow us to correctly weight the influence of the detector in the overall FR results. Finally, recognizability  would allow proper weighting of different samples in set-based face recognition. 

So, recognizability and the addition of an unrecognizable identity (UI) class would address the issues set forth in the introduction. But {\em what is the unrecognizable identity?} Is it an actual identity, or just a moniker for the interstitial space around the decision boundaries among all known identities?

If we represent each face via an embedding in a compact metric space, such as the hypersphere, perturbing the image until it is unrecognizable moves the corresponding embedding close to a decision boundary. So, it is reasonable to expect that the embeddings of UIs distribute along the boundaries of decision regions with no particular relation to each other:  Distant identities, when perturbed, would become distant UIs. Instead, we observe the following phenomenon: {\em When training a FR system without any UIs, and using the resulting embedding to cluster identities (both recognizable and not), UIs cluster together in representation space}, despite being unrecognizable versions of different identities that may otherwise be far in representation space. 

This phenomenon, illustrated in Fig.~\ref{fig_tsne_trajectory}, is counter-intuitive at many levels: First, UIs do not distribute near the boundary between different identities, but rather close to each other and far from the corresponding identities. This happens without imposing any loss on the distance among UIs -- for they are not even included in the training set for the FR -- and likely made possible by the geometry of the high-dimensional hypersphere.\footnote{This phenomenon is unrelated to the known collapse of representations of low-quality images towards the origin of a linear embedding space \cite{o2018face}, as here the embedding is constrained to be on the hypersphere.} Second, this phenomenon is not only due ``low-quality'' images of UI being {\em visually similar to each other}. Unlike other domains where motion-blurred images form their own cluster, distinct from the low-resolution cluster, here the UI cluster is {\em highly heterogeneous,} with images that exhibit occlusion, optical or motion blur, low resolution, poor illumination, etc. So, the clustering of UIs is not simply due to visual similarity. Sample of these phenomena are illustrated in Fig.~\ref{face_clusers} and Sec.~\ref{RS}. We conjecture that this behavior is specific to FR,  which is a fine-grained categorization task, where nuisance variability can cause coarse-grained perturbations that move the corresponding samples out-of-domain. 

We are now ready to tackle the main goal: Leverage the key empirical observation above to {\em harness unrecognizable faces to improve face recognition.}

\subsection{Contributions and Related Work}

Our contribution is three-fold, corresponding to the three questions posed in the introduction: 

1) We propose a measure of recognizability that leverages the existence of a single UI cluster in the learned embedding.  The  {\em ``embedding recognizability score''} (ERS) is simply the Euclidean (cordal) distance of the embedding of an image from the center of the UI cluster in the hypersphere.

2) We use the ERS to mitigate the detrimental effect of using different datasets for FD and FR. This results in 58\% error reduction (at FAR=1e-5 on the IJB-C Covariate Verification benchmark) in single-image face recognition, without impacting the performance of the detector.

3) We propose an aggregation method for set-based face recognition, by simply calculating the weighted average relative to the ERS, and report 24\% error reduction (at FAR=1e-5 on IJB-C) compared with uniform averaging.  

We emphasize that unrecognizable faces are still included in the evaluation and the improvement is due to the proposed matching method using ERS (Sec.~\ref{ERS_single_FR}).

The face recognition literature is gargantuan; by necessity, we limit our review to the most closely related methods, cognizant that we may omit otherwise important work. Like most, we use a deep neural network (DNN) model to compute our embedding function mapping an image $x$ and a bounding box $b$ to a (pseudo-)posterior score $\phi(x_b) \propto \log P(y | x, b)$ where $y \in \{1, \dots, K\}$ denotes the identity for the case of search, and $K = 2$ for verification \cite{deng2020sub,Deng2018ArcFaceAA,Liu2017SphereFaceDH,wen2016discriminative,Parkhi2015DeepFR,Schroff2015FaceNetAU,Lu2019UnsupervisedDD,Wang2018DeepFR,zhao2018towards,Lu2018DeepCR,Wang2016StudyingVL,Jian2015SimultaneousHA}. 
Our work relates to efforts to understand the effect of image quality on face recognition: Some explicitly modeled image quality in face aggregation~\cite{gong2019video,xie2018comparator,xie2018multicolumn,yang2017neural,liu2017quality}, others utilized correlation among images from the same person to improve set-based~\cite{liu2019permutation,liu2018dependency} or video-based~\cite{rao2017learning} face recognition. More recently, \cite{hernandez2020biometric,hernandez2019faceqnet} developed face quality assessment tools and methods~\cite{terhorst2020ser}, probabilistic representation to model data uncertainty~\cite{shi2020towards,chang2020data,shi2019probabilistic}, as well as explicit handling of quality variations with sub-embeddings~\cite{shi2020towards}. 

\subsection{Implications on Bias and Fairness}

As with any data-driven algorithmic method, a trained FR system is subject to statistical bias due to the distribution of the training set, and more general algorithmic bias engendered by the choice of models, inference criteria, optimization method, and interface. In addition to those externalities, there can be intrinsic sources of bias that exist before any algorithm is implemented or dataset collected: Light interacts with matter in a way that depends on sub-surface scattering properties of materials, which is different depending on skin color and affects recognizability by {\em any} system, natural or artificial. 
Moreover, clothing, hairdo, makeup, and accessories similarly can impact recognizability.
We therefore expect that the UI will {\em not} represent an unbiased sampling of the population and reflect physical and phenomenological characteristics of the scene, the illuminant, the sensor, and the capture method,  regardless of the algorithm used to infer recognizability. {\em In addition}, recognizability is based on statistical properties of the embedding, which is subject to the usual data and algorithmic biases. 
It is not clear how to balance different populations in the UI.
We defer these important and delicate questions to the many studies on bias and fairness, and focus on the orthogonal question of how to deal with recognizability irrespective of what algorithm or system is used for recognition.

\section{Face recognizability in face recognition}\label{UI_in_FR}
To consider face recognizability in the context of face recognition systems, we first study the impact of face recognizability degradation to FR accuracy and reveal why it is important to account for input data recognizability. 
Then we propose a measure of face recognizability w.r.t. the face embedding model. 
By factoring the recognizability measurement into FR prediction decisions, we propose methods to mitigate the detriment of unrecognizable faces to FR systems.

\subsection{Observing Recognizability}

The observation that {\bf UIs cluster together} can be illustrated using face embeddings~\cite{wang2018cosface} 
on images of $8$ randomly sampled celebrities from the IJB-C dataset. 
We synthesize UI images by perturbing images with increasing Gaussian blur, motion blur, occlusions, etc.
The t-SNE visualization of the embeddings is shown in Fig.~\ref{fig_tsne_trajectory}, where recognizable faces form separate clusters corresponding to their identities, but as they become increasingly unrecognizable they do not distribute around the boundary between identities, nor around their centroid, but rather around a distinct cluster, the UI cluster. 

\begin{figure}[!t]
\centering
\includegraphics[width=\linewidth]{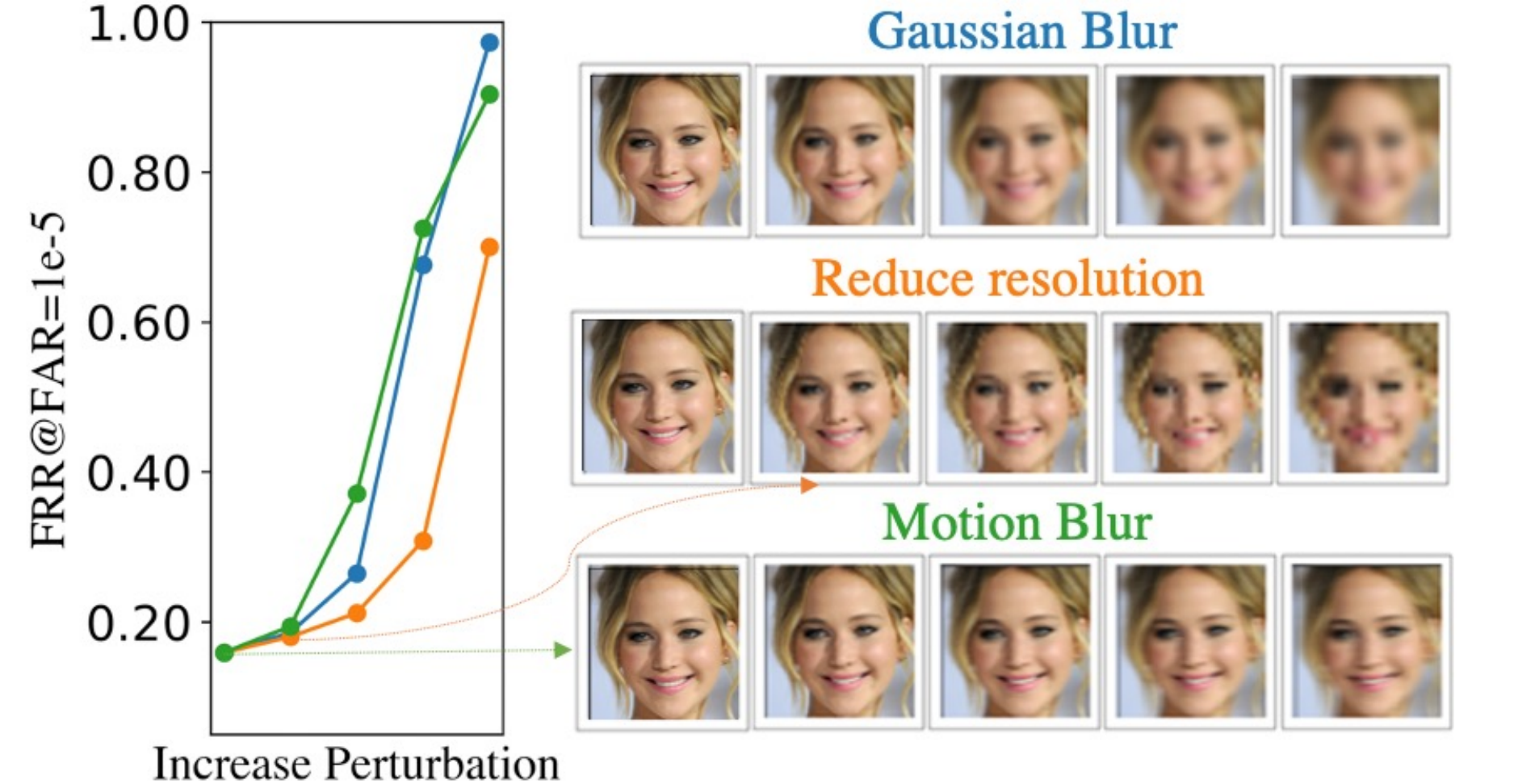}
\caption{\label{fig_arti_noise} Face verification accuracy can be heavily impacted by image perturbation, evidenced by IJB-C Template-based Face Verification benchmark results: left x-axis marks perturbation level low to high and y-axis marks the FRR@FAR=1e-5, in correspondence with images on the right. Perturbation types are color-coded.}
\end{figure}

To validate this result, we perform large-scale face clustering~\cite{day1984efficient} 
on datasets where there are artificial or natural low recognizability (LR) images of faces or non-faces.
To exemplify artificial LR images, we take a subset of 10k faces from DeepGlint~\cite{deepglint}, a face embedding training dataset. We randomly apply Gaussian blur, motion blur, resolution reduction, rotation, affine transformation, or occlusion to corrupt $10\%$ (this guarantees the UI cluster to be larger than any of the id classes) of the faces. 
We refer to this as the Perturbed DeepGlint Subset.
After running face clustering with the HAC algorithm~\cite{day1984efficient} on the normalized features, a UI cluster emerges. 
The cluster size is larger than all the other clusters by more than two orders of magnitude.
For natural LR images, we run clustering on face crops detected by a state-of-the-art FD model on the WIDERFace~\cite{yang2016wider} dataset.
Again, we obtain a heterogeneous cluster composing of solely unrecognizable and visually dissimilar faces. 
The identity-agnostic UI images distribute closely in one cluster and away from identity-based clusters. 
This explain two typical types of face quality-related errors in FR reported in several prior works~\cite{chang2020data,shi2019probabilistic}: false positive matches between low-quality faces of different ids and false negatives between high and low-quality faces of the same id.

The clustering results on face datasets are surprising as low-quality images (for instance, a quality degraded ImageNet database) ordinarily tend to cluster by appearance: blurred images form a cluster which is distinct from that of dark images, and that of motion-blurred images, etc. 
We conjecture that the existence of UI has to do with the fine-grained nature of the face domain because we discover similar phenomena also emerge in other fine-grained recognition tasks like person re-identification and fashion retrieval.

What impact could the UIs have on the downstream FR task?
To quantify this, we add recognizability corruption (Gaussian blur, reduced resolution, and linear motion blur) to images in the IJB-C face verification benchmark and show the change of error rates in Fig.~\ref{fig_arti_noise}. 
We observe a clear trend of error increase as more corruption is added.
This shows the risk of not considering the recognizability of the images in recognition and calls for a mitigation solution.
Luckily, the fact that \emph{the UI is distant from recognizable identities in the embedding space},
also suggests a way of {\em measuring recognizability by measuring their distances to the UI} and factor recognizability into face recognition prediction, as we detail in Sec.~\ref{RS}.

\subsection{Accounting for Recognizability}
Based on the hypothesis that distance to the UI cluster centroid could serve as a measure of the face recognizability, we use the distance as an \emph{embedding recognizability score} (ERS), which requires no additional training nor annotation, and can be easily incorporated into both single-image and set-based face recognition.  

\subsubsection{The Embedding Recognizability Score} \label{RS}
\begin{figure}[!t]
\centering
\includegraphics[width=0.98\linewidth]{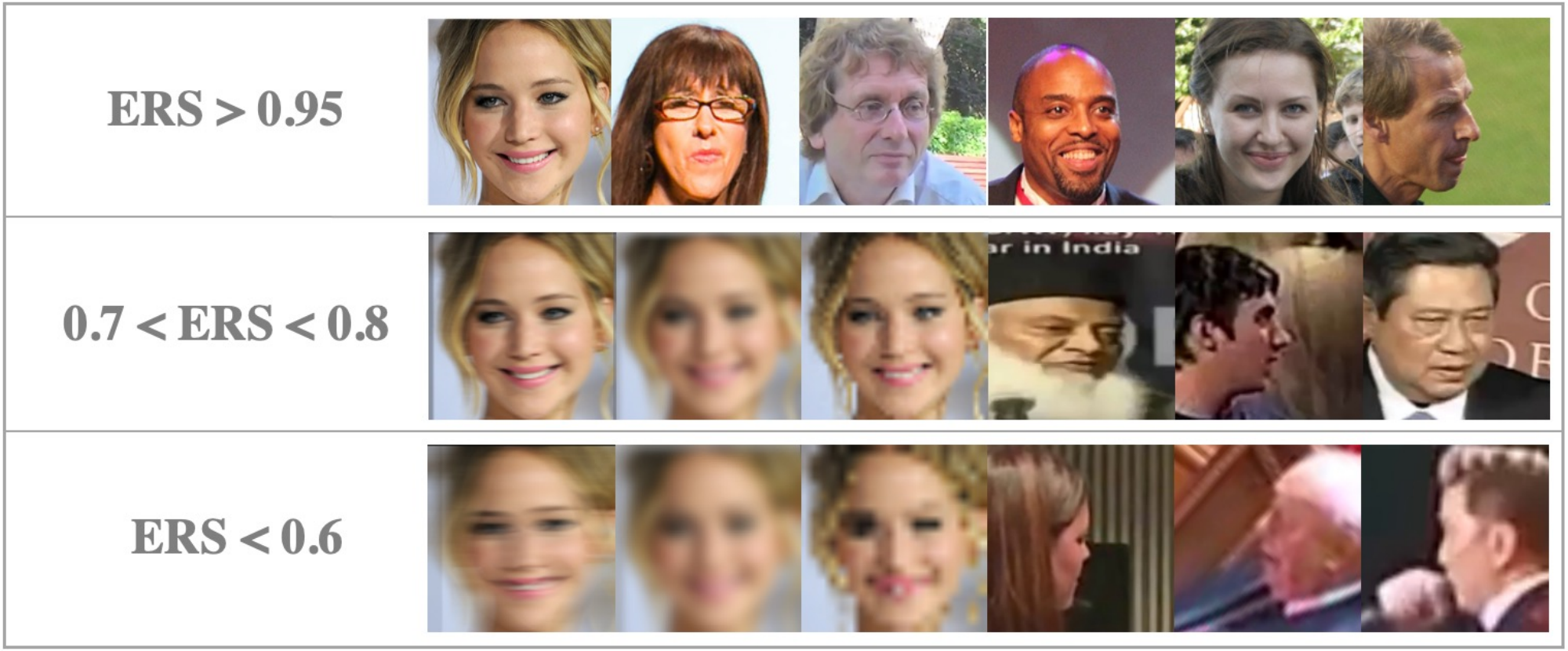}
\caption{\label{fig_RS_rank} Sample images from IJB-C dataset, grouped by embedding based recognizability scores (ERS): high ($>$0.95), middle (0.7$-$0.8), and low ($<$0.6). Images with low ERS are hard to recognize even for human viewers.}
\end{figure}

We define Embedding Recognizability Score (ERS) to be the distance between an embedding vector and the average embedding of UI images.  
These UI images can be obtained by either clustering artificial degraded training data of the embedding model or large-scale in-the-wild face detection datasets such as WIDERFace~\cite{yang2016wider} and taking the resultant large cluster.  
The normalized average embedding of UI images $\mathbf{f}_{UI}$, which we call the UI centroid (UIC), is then used to represent the UI.
Given an embedding vector $\mathbf{f}_i$, its ERS $e_i$ is defined as 
\begin{align}
    e_i = 1 - \langle\mathbf{f}_{UI}, \mathbf{f}_i\rangle.
\end{align}
We illustrate the correlation between ERS and recognizability in Fig.~\ref{fig_RS_rank} by taking images from the IJB-C dataset and grouping them by their ERS scores. 
The ERS decrease is accompanied by face quality variations such as occlusion distortion, larger pose, and increased image blurriness.

\subsubsection{ERS in Single Image-Based Recognition}\label{ERS_single_FR}

\noindent\textbf{Face verification} aims to determine whether two face images, $x_1$ and $x_2$, belong to the same person. They have corresponding groundtruth identity labels $y_1$ and $y_2$.
The face embedding model represents an input face image $x_i$ as the feature vector $\mathbf{f}_i \in \mathcal{R}^d$. 
Without considering recognizability, the estimated probability of two images being from the same person is usually
\begin{align}
    p(\hat{y}_1 = \hat{y}_2 | x_1, x_2) = s(\mathbf{f}_1, \mathbf{f}_2).
\end{align}
Here $s(\cdot, \cdot)$ is the cosine similarity function, and $\hat{y}_i$ is the estimated identity label of $x_i$, which in practice not explicitly computed. 
Then a binary prediction is made based on whether the probability is higher than a empirically set threshold $\tau$. 

To take ERS into account, we allow the system to predict ``unsure'' instead of ``same'' or ``different'' when either $e_1$ or $e_2$ is below a threshold $\gamma$. 
When the ERS of $x_1$ or $x_2$ is low, we observe the recognizability to be low. 
The similarity between them cannot be reliably determined to predict if they belong to the same id. 
In applications, one can choose further actions on the unsure cases. 
For example, in our experiments on the IJB-C Covariate Test, we choose to predict all unsure cases as not belonging to the same person due to the empirical risk of false matches.

\noindent\textbf{Face identification} aims to tell whether one query image $x_i$, represented as $\mathbf{f}_i$ belongs to one of $N$ indexed identities in the gallery, represented as $\{ \mathbf{g}_j\}_{j=1}^N$. 
\footnote{In this work we only deal with the ``open-set'' setting~\cite{maze2018iarpa} where predicting a query has no match in the gallery is allowed, as it is the most common usage in real-world FR systems.}
Here we assume that the gallery has mostly recognizable images.
Without considering recognizability, the decision function $S(\mathcal{R}^d) \rightarrow [0, \ldots, N]$ is
\begin{align}
    S(\mathbf{f}_i ; \{\mathbf{g}_j\}) = \mathbbm{1}[\max_j s(\mathbf{f}_i, \mathbf{g}_j) \ge \tau] \cdot \argmax_{j = 1, \ldots, N} s(\mathbf{f}_i, \mathbf{g}_j).
\end{align}
The query has a positive match when the maximal similarity score to any of the gallery images is above the threshold $\tau$. 
With ERS, the decision function becomes
\begin{align}\label{eq:search_w_ers}
    S^\prime(\mathbf{f}_i ; \{\mathbf{g}_j\}) = S(\mathbf{f}_i ; \{\mathbf{g}_j\})\mathbbm{1}(e_i \ge \gamma).
\end{align}
Again, when $e_i$ is lower than $\gamma$, it becomes an unsure case, we predict there is no positive match. 
If there is no guarantee that the gallery images are mostly recognizable, ERS could be applied to reject retrieved gallery images below $\gamma$.

\subsubsection{ERS in Image Set-based Face Recognition}\label{ERS_set_FR}
In set-based face recognition, we have prior knowledge that each set or template~\cite{maze2018iarpa}, contains one or multiple face images belonging to a single person. 
Set-based face recognition also usually consists of face verification and face identification. 
We first extract feature vectors using the embedding model for every image in each set $\theta_i$ containing images $\{x_i^l\}$ as $\{\mathbf{f}_i^l\}_{l=1}^{|\theta_i|}$, where $|\theta_i|$ is the cardinality of $\theta_i$. 
Then the feature vectors are aggregated into one feature vector $\mathbf{f}_i$.
After aggregation, the processing is the same as in the single image case. 
We design an aggregation function weighted by the ERS of each image as
\begin{align}\label{eq:aggregation}
    \mathbf{f}_i = \sum_{l=1}^{|\theta_i|} \frac{w(e_i^l) \mathbf{f}_i^l}{\sum_l e_i^l}, 
    \quad e_i = \frac{\sum_l e_i^l}{|\theta_i|},
\end{align}
where $\mathbf{f}_i$ and $e_i$ denote the aggregated feature vector and ERS for the set $\theta_i$, $w:\mathcal{R} \rightarrow \mathcal{R}^+$ denotes the weighting function based on ERS.
The aggregated feature vectors can then be used as in the single image cases.  
We discuss choices of weighting function in the ablation study.


\section{Experiments}\label{sec:experiments}
We examine the effectiveness of ERS in face recognition on multiple benchmarks.
We consider two face image quality assessment methods as the baselines: FaceQnet~\cite{hernandez2019faceqnet} and SER-FIQ~\cite{terhorst2020ser}.
In set-based face recognition, we compare the ERS based aggregation function with other set-based recognition methods: NAN~\cite{yang2017neural}, Multicolumn\cite{xie2018multicolumn},  DCN~\cite{xie2018comparator}, PFE~\cite{shi2019probabilistic} and DUL~\cite{chang2020data}.

\noindent\textbf{Implementation details}
We use the deep learning framework MXNet~\cite{chen2015mxnet} in our model training and evaluation. 
We train a face embedding model using CosFace~\cite{wang2018cosface} loss, ResNet-101 (R101)~\cite{he2016deep}  backbone
and DeepGlint-Face dataset (including MS1M-DeepGlint and Asian-DeepGlint)~\cite{deepglint}. 
HAC algorithm~\cite{day1984efficient} is used to cluster extracted embeddings and generate UI clusters.
We select threshold $\gamma=0.60$ for ERS via cross-validation on the TinyFace~\cite{wang2014low} benchmark. 

\begin{figure}[!t]
\begin{center}
\includegraphics[width=\linewidth]{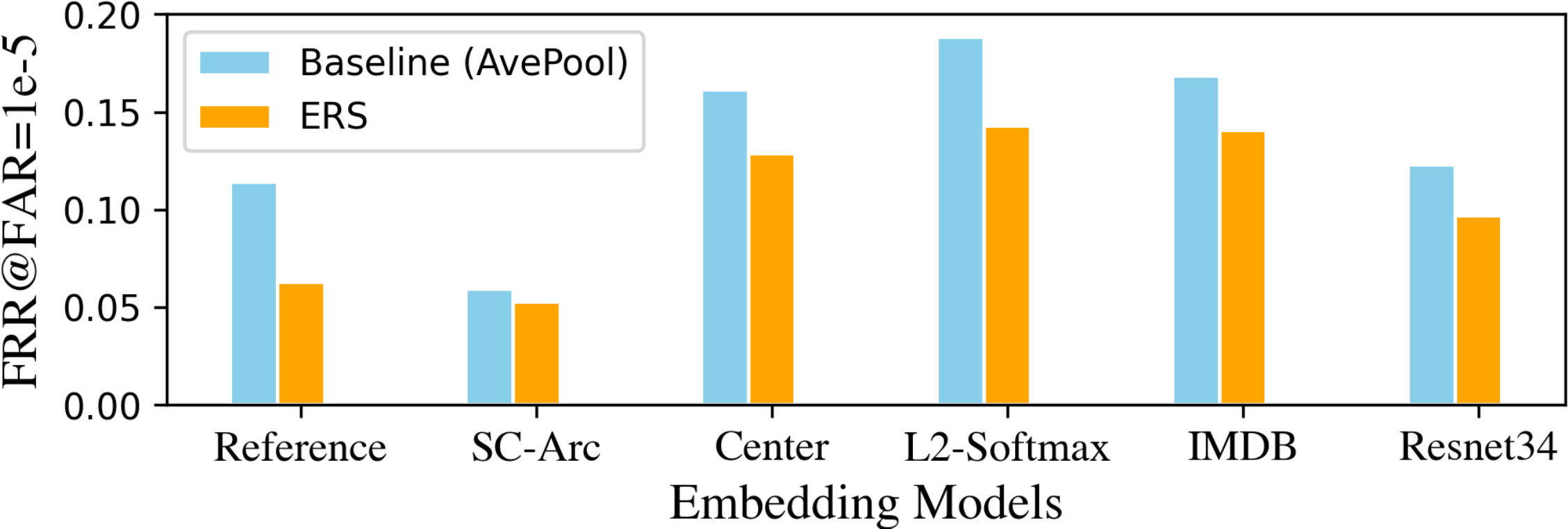}
\end{center}
   \caption{\label{fig_ablation_settings} Pairwise performance comparison between baseline (average pooling) and our method (ERS) on IJB-C template across different settings, our method achieves consistent error reduction. Training settings: Reference (R101, CosFace, DeepGlint), SC-Arc(R101, Sub-center Arcface, DeepGlint), Center (R101, Softmax+Center, DeepGlint), L2-Softmax (R101, $\ell_2$-Softmax, DeepGlint), IMDB (R101, CosFace, IMDB), ResNet34 (R34, CosFace Loss DeepGlint).}
\end{figure}

\noindent\textbf{Evaluation data}
Single-image and set based face recognition experiments are run on the IARPA Janus Benchmark-C (IJB-C)~\cite{maze2018iarpa}
\footnote{This paper contains or makes use of the following data made available by the Intelligence Advanced Research Projects Activity (IARPA): Benchmark C (IJB-C) data detailed at Face Challenges homepage. For more information see https://nigos.nist.gov/datasets/ijbc/request.}.
We introduce the IARPA Janus Benchmark-C (IJB-C)~\cite{maze2018iarpa} to run evaluations.
IJB-C dataset suite contains in-the-wild celebrity media, including photos and videos, and multiple predefined benchmark protocols.
We introduce four protocols from IJB-C: 
1) \emph{IJB-C Covariate Face Verification Test} for single image-based face verification, 2) \emph{IJB-C Template-based Face Verification} for set-based face verification, 3) \emph{IJB-C Template-based Face Search} for set-based face search, and 4) \emph{IJB-C Test10: Wild Probe with Full Motion Video Face Search} for set-based video face search.
It is worth noting that the IJB-C Test10 contains video frames with strong motion blur, forming an FR testbed with unrecognizable video faces.

\noindent\textbf{Evaluation Metric}
Methods performance are measured on two standard face recognition test cases: 1:1 verification and 1:N search. 
For face verification, we measure False Reject Rates (FRRs, also known as $1-$ True Acceptance Rate), at a set of False Acceptance Rates (FARs).
For face search, we measure False Negative Identification Rates (FNIRs, or $1-$ True Positive Identification Rate) at different False Positive Identification Rates (FPIR), as well as top-$K$ accuracy.

\subsection{ERS in Single Image-Based Face Verification }

\begin{table}[!t]
\centering
\begin{adjustbox}{max width=\linewidth}
\begin{tabular}{c|c|c|c|c} 
\hline
\multirow{2}{*}{Mehod}   & \multicolumn{4}{c}{Verification FRR@FAR}                           \\ 
\cline{2-5}
                         & 1e-6            & 1e-5            & 1e-4            & 1e-3             \\ 
\hline
Baseline                 & 0.6976          & 0.4540          & 0.1747          & 0.0714           \\ 
\hline
FaceQnet~\cite{hernandez2019faceqnet} ($\gamma=1.00$) & 0.6976          & 0.4540          & 0.1747          & 0.0714           \\ 
\hline
SER-FIQ~\cite{terhorst2020ser} ($\gamma=0.83$) & 0.4027          & 0.2023          & 0.1164          & 0.0717           \\ 
\hline
ERS ($\gamma=0.60$)      & \textbf{0.3819} & \textbf{0.1885} & \textbf{0.1113} & \textbf{0.0673}  \\
\hline
\end{tabular}
\end{adjustbox}
\caption{\label{table_covariate_benchmark} Comparison of recognizability conditioned face verification on IJBC Covariate Verification benchmark. We compare with FaceQnet~\cite{hernandez2019faceqnet} and
and SER-FIQ~\cite{terhorst2020ser} as the alternatives for face recognizability measure. (As best thresholds are not suggested from the other works, we densely tested multiple thresholds and select their best results for comparison.)}
\end{table}

The IJB-C Covariate Test protocol benchmarks single-image based face verification. 
In Table~\ref{table_covariate_benchmark}, we evaluate our approach on this benchmark, and compare with FaceQnet~\cite{hernandez2019faceqnet} and SER-FIQ~\cite{terhorst2020ser}
as the alternatives for face recognizability measure.
With the same embedding model, using ERS as the recognizability measure leads to $58\%$ error reduction comparing to that of not considering recognizability. 
The alternative recognizability measure FaceQnet cannot improve the baseline at all thresholds, and SER-FIQ is able to help reduce errors, but with less effect.

\begin{table*}
\centering
\begin{adjustbox}{max width=\textwidth}
\begin{tabular}{c|c|c|c|c|c|c} 
\hline
\multirow{2}{*}{Method} & \multirow{2}{*}{Backbone}                                                                                  & \multirow{2}{*}{Train Data} & \multicolumn{4}{c}{IJB-C Verification FRR@FAR}                                                                                                                                                \\ 
\cline{4-7}
                        &                                                                                                            &                             & 1e-5                                          & 1e-4                                          & 1e-3                                          & 1e-2                                           \\ 
\hline

Multicolumn~\cite{xie2018multicolumn}             & ResNet50                                                                                                   & VGGFace2                    & 0.2290                                        & 0.1380                                        & 0.0730                                        & 0.0320                                         \\ 
\hline
DCN~\cite{xie2018comparator}                     & ResNet50                                                                                                   & VGGFace2                    & -                                             & 0.1150                                        & 0.0530                                        & 0.0170                                         \\ 
\hline
PFE~\cite{shi2019probabilistic}                     & 64-Layer CNN                                                                                               & MS-Celeb-1M                 & 0.1036                                        & 0.0675                                        & 0.0451                                        & 0.0283                                         \\ 
\hline
DUL~\cite{chang2020data}                     & ResNet64                                                                                                   & MS-Celeb-1M                 & 0.0977                                        & \textbf{0.0539}                               & 0.0530                                        & -                                              \\ 
\hline
NAN~\cite{yang2017neural}                     & ResNet64 trained with Cosface Loss                                                                                                   & DeepGlint                   & 0.1023                                        & 0.0582                                        & 0.0347                                        & 0.0206                                         \\ 
\hline
Baseline (AvePool)                 & \multirow{2}{*}{ResNet64 trained with Cosface Loss}                                                        & \multirow{2}{*}{DeepGlint}  & 0.1262          & 0.0639          & 0.0332          & 0.0165           \\
ERS                     &                                                                                                            &                             & \textbf{0.0929} & 0.0547          & \textbf{0.0312}  & \textbf{0.0157}   \\ 
\hline
\hline
Baseline (AvePool)                 & \multirow{2}{*}{\begin{tabular}[c]{@{}c@{}}ResNet101 trained with~\\Sub-Center Arcface Loss \end{tabular}} & \multirow{2}{*}{DeepGlint}   & 0.0592          & 0.0406          & 0.0271         & \textbf{0.0157}            \\
ERS                     &                                                                                                            &                             & \textbf{0.0528} & \textbf{0.0363} & \textbf{0.0245}  & 0.0159  \\
\hline
\end{tabular}
\end{adjustbox}
\caption{\label{table_ijbc_sota} Benchmark results on IJB-C template-based verification. On top, we compare using ERS for aggregation with the baseline of averaging embedding (AvePool) and other face aggregation methods. The results are reported at different FAR levels. We report error rates to better illustrate the difference between methods.
ERS is also effective on the latest embedding model (Sub-Center ArcFace).
}
\end{table*}

\subsection{ERS in Set-Based Face Recognition} \label{face benchmark} 
We evaluate set-based face recognition using ERS as described in Sec.~\ref{ERS_set_FR}.  
Results on the IJB-C Template-Based Face Verification and Search are illustrated in Table~\ref{table_ijbc_sota} (top). 
We adapt media-pooling~\cite{crosswhite2018template} in set-based benchmarks.
We compare ERS based method to the baseline of simple averaging without considering recognizability (AvePool). 
We also compare our approach with other methods developed for set-based face recognition.
It can be seen that our approach significantly reduces recognition errors of the baseline.
Our results are comparable or better than other complex methods developed for set-based recognition. 

We also evaluate ERS on the IJB-C Test10 benchmark to test our algorithms on videos in-the-wild and implement an average pooling baseline and learning-based method NAN~\cite{yang2017neural} for comparison.
The results are summarized in Table~\ref{table_test_10_self}.
It can be seen that although obtained from images, ERS is also able to improve recognition accuracy in challenging video data.

\noindent\textbf{Improving state-of-the-art face embedding models}.
In Fig.~\ref{fig_tsne_trajectory} we illustrate the UI cluster using the Cosface~\cite{wang2018cosface} for embedding. 
The UI clustering phenomenon is not limited to that embedding model.
We empirically find that multiple other face embedding models have similar behaviors, despite their different loss functions, size of training datasets, and backbone architectures.
We conjecture that the existence of UI clusters may be attributed to the nature of face recognition as a fine-grained categorization task. 
Since ERS is easy to obtain for any face embedding model without extra training or annotation, we test applying ERS to multiple state-of-the-art face embedding models trained under different settings, including
(1) loss function designs: Sub-center ArcFace \cite{deng2020sub}, Softmax+Center Loss \cite{wen2016discriminative} and $\ell_2$-Softmax Loss \cite{ranjan2017l2} ; 
(2) training dataset: IMDB \cite{wang2018devil} and DeepGlintFace; 
(3) backbone architectures: ResNet-101, ResNet-50 and ResNet-34 \cite{he2016deep}.

Results on the IJB-C face verification benchmark are illustrated in Fig.~\ref{fig_ablation_settings}.
We observe consistent error reduction on all tested models, including 10\% error reduction at FAR=1e-5 on strong baseline from Sub-center ArcFace \cite{deng2020sub}. Full results can be found in Table~\ref{table_ijbc_sota} (bottom).
This suggests that the ERS can be an easy plug-in to existing face recognition systems to reduce recognition errors. 

\begin{table}[!t]
\centering
\begin{adjustbox}{max width=\linewidth}
\begin{tabular}{c|c|c|c|c|c|c} 
\hline
\multirow{2}{*}{Method} & \multicolumn{4}{c|}{Identification FNIR@FPIR}                                                                                                                                                                         & \multicolumn{2}{c}{Rank-N Error}     \\ 
\cline{2-7}
                        & 1e-4                                                & 1e-3                                                & 1e-2                                                & 1e-1                                                & 1                & 5                 \\ 
\hline
AvePool      & 0.8607                                              & 0.6840                                              & 0.4129          & 0.2029          & 0.1564           & 0.1137            \\ \hline
NAN    & 0.8566          & 0.6697          & 0.3726          & 0.1956          & 0.1518           & 0.1100            \\\hline
ERS                     & \textbf{0.8299} & \textbf{0.6096} & \textbf{0.3054} & \textbf{0.1807} & \textbf{0.1457}  & \textbf{0.1055}   \\\hline
\end{tabular}
\end{adjustbox}
\caption{\label{table_test_10_self} IJB-C Test 10: Wild Probe with Full Motion Video Face Search results, tested using backbone ResNet101 trained on DeepGlintFace with CosFace~\cite{wang2018cosface} loss. In comparison with baseline average pooling and NAN~\cite{yang2017neural} trained on top of the same backbone model, our ERS-based aggregation achieves the best performance.}
\end{table}


\subsection{Ablation study}

\noindent\textbf{Generation of the UI centroid}. \label{abl_generation_UI}
The UI centroid (UIC) is a key component to our ERS-based methods.
Based on the observation that heterogeneous UI images cluster in one, we hypothesize that it is possible to obtain UIC from different data distributions where there are low recognizability images.
Accordingly, our methods should be robust to the dataset choices.
We compare UIC generated from the Perturbed DeepGlint Subset (PDS) and face detection datasets WIDERFace and FDDB~\cite{jain2010fddb}.
In table~\ref{table_ab_NE}, we list the average cosine distances of each UI cluster, a comparison between generated UICs, and associated aggregation results.
It can be seen that ERS is not sensitive to the sources dataset.

When clustering across different embedding models, we find resultant images in the UI clusters to have significant overlap. 
Further experiments prove a fixed set of UI images generated from one embedding model can be reused by other models for obtaining UIC.
In fact, we obtain a UI set from a ResNet101-based Cosface model, and reuse it to generate UIC for all other models in Fig.~\ref{fig_ablation_settings} and Table~\ref{table_test_10_self}.
This way, the application of our method can be simplified by skipping face clustering once we have a UI image set.

\begin{table}[!t]
\centering
\begin{adjustbox}{max width=\linewidth}
\begin{tabular}{c|c|c|c}
\hline
Clustering dataset            & PDS    & WIDERFace & FDDB       \\ \hline
UI cluster average distance   & 0.3907 & 0.3213    & 0.4344    \\ \hline
UIC distance to that from PDS & 0.0000 & 0.04736    & 0.0395    \\ \hline
w/ ERS FRR@FAR=1e-5       & 0.0623 & 0.0627    & 0.0625   \\ \hline
\end{tabular}
\end{adjustbox}
\caption{\label{table_ab_NE} Comparison between UI generated from different datasets and ERS results on IJB-C Template-based Face Verification. Without ERS the FRR@FAR=1e-5 is 0.1140.}
\end{table}

\noindent\textbf{Choice of the weighting function $w$}. 
Using ERS in set-based face recognition requires a choice of the weighting function $w$.
We compare different choices of $w$, including identity, exponential, and square on the IJB-C Template-Based Face Verification benchmark.
We also compare with two special choices that average the images with top-1\% and top-10\% ERSs within a set.
From table~\ref{table_ab_scoring} we can see square function achieves the best results. We use it in other experiments without special notes.

\begin{table}[!htb]
\centering
\begin{adjustbox}{max width=\linewidth}
\begin{tabular}{c|c|c|c|c|c} 
\hline
$w$   & $e_i$  & $softmax(e_i)$  & $e_i^2$  & $e_i$  & $e_i$     \\ 
\hline
Set Selection    & N/A    & N/A             & N/A      & top 1  & top 10\%  \\ 
\hline
FRR@FAR 1e-5 & 0.0684 & 0.1393          & \textbf{0.0627}   & 0.2196 & 0.2025    \\
\hline
\end{tabular}
\end{adjustbox}
\caption{\label{table_ab_scoring} Comparison between different ERS-based aggregation methods on  IJBC Templated-based Face Verification benchmark, basenet ResNet101 trained with Cosface~\cite{wang2018cosface}.}
\end{table}

\subsection{Exploration on increasing ERS} \label{FD}
Higher ERSs usually associate with faces with better recognizability.
It is interesting to explore whether we can increase the ERS of a given image. 
We present the results of a naive method for this: enhance the face feature by removing its projection on the direction of the UI representation to yield high ERS ($e=1$) features.
Formally, we take the raw feature embedding $\mathbf{f}$, unit vector $\mathbf{f}_{UI}$ and calculate the feature
\begin{align}
\mathbf{v}^{id}=\mathbf{f}-\langle\mathbf{f},\mathbf{f}_{UI}\rangle\mathbf{f}_{UI}.
\end{align}
After $\ell_2$ normalization we get the ERS-enhanced features $\mathbf{f}^{id}$.
From table~\ref{table_ERS_enhanced} we can see average pooling with the ERS-enhanced features surpasses the baseline by a large margin and achieves comparable results to ERS weighted aggregation. This suggests increasing embedding ERS can make a meaningful difference in benchmark results, achieving higher ERS through more advanced techniques may lead to a further increase in recognition accuracy. 

\begin{table}[!t]
\centering
\begin{adjustbox}{max width=\linewidth}
\begin{tabular}{c|c|c|c|c} 
\hline
\multirow{2}{*}{}      & \multicolumn{4}{c}{FRR@FAR}                                                                                                                                                               \\ 
\cline{2-5}
                      & 1e-5                                          & 1e-4                                          & 1e-3                                          & 1e-2                                           \\ 
\hline
Baseline AvePool     & 0.1140          & 0.0468          & 0.0264          & 0.0141           \\ 
\hline
ERS (WeightedPool)     & \textbf{0.0627} & \textbf{0.0393} & \textbf{0.0243} & \textbf{0.0140}  \\ 
\hline
ERS-Enhanced AvePool & 0.0644          & 0.0410          & 0.0255          & 0.0144          \\
\hline
\end{tabular}
\end{adjustbox}
\caption{\label{table_ERS_enhanced} Comparison among average pooling, ERS weighted pooling and average pooling with ERS enhanced features on IJB-C Template-based Face Verification benchmark, basenet is R101 trained with Cosface loss.}
\end{table}


\subsection{Exploration on other vision tasks}\label{other_task}

\begin{figure}[!t]
\begin{center}
\includegraphics[width=\linewidth]{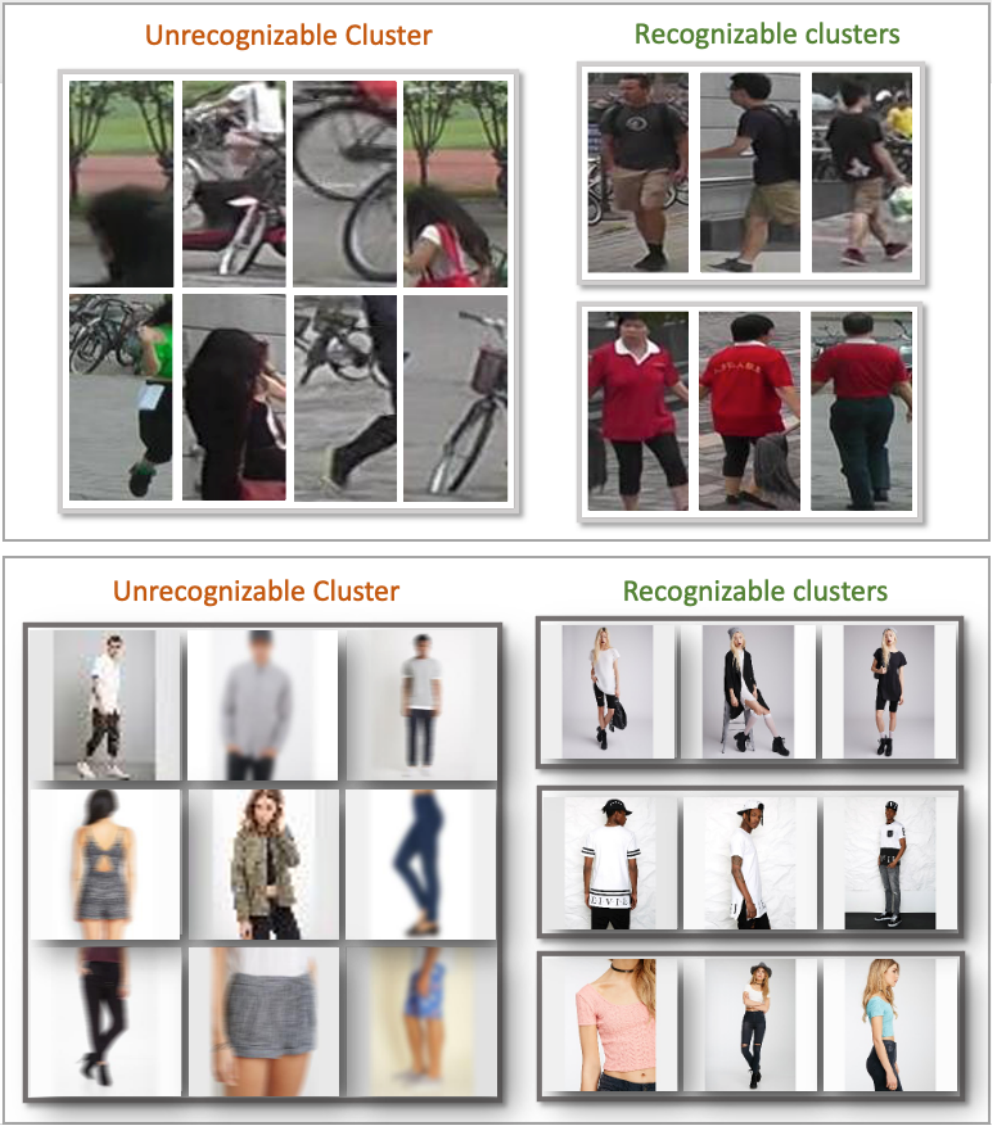}
\end{center}
  \caption{\label{fig_reid_clustering} Similar to our findings in face clustering, miscellaneous low recognizability examples can gather in one cluster in person re-id (top, using Market1501~\cite{zheng2015scalable} dataset) and fashion retrieval datasets (bottom, using Deepfashion~\cite{liuLQWTcvpr16DeepFashion} dataset). 
  }
\end{figure}

\begin{figure}[!t]
 	\centering
 	\includegraphics[width=0.98\linewidth]{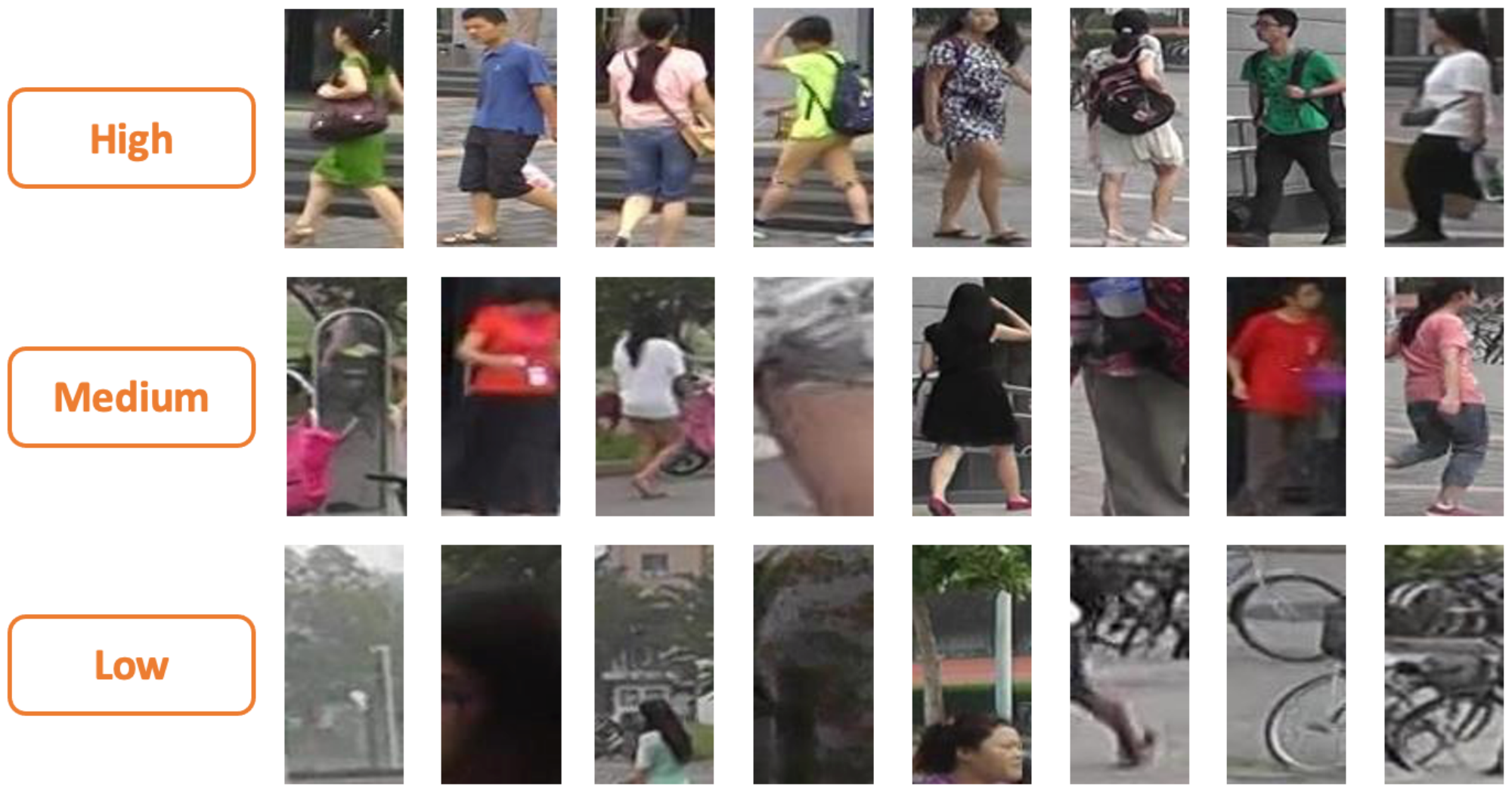} 
 	\caption{\label{fig_reid_RS_rank} Images from Market1501 grouped by high, medium, low ERS scores. Positive correlation between recognizability and ERS can be observed.}
\end{figure}

\begin{figure}[t!]
	\centering
	\includegraphics[width=0.98\linewidth]{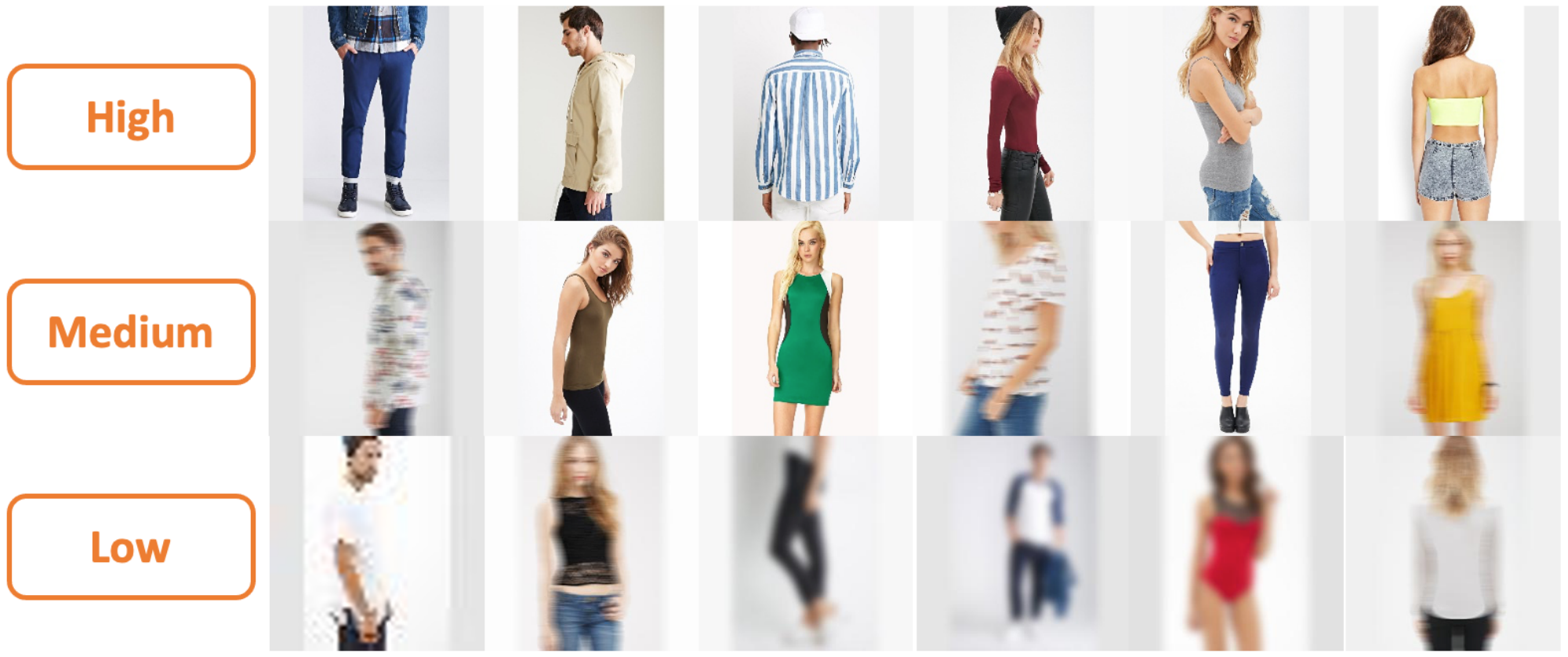}
	\caption{\label{fig_inshop_RS_rank} Images from Deepfashion with synthetic corruption grouped by high, medium, low ERS scores. Positive correlation between recognizability and ERS can be observed.}
\end{figure}

As an extended study, we explore clustering on person re-identification (re-id) and image retrieval to see whether the low recognizability clustering phenomenon exists and our method can be applied accordingly.
We perform re-id embedding clustering on Market1501~\cite{zheng2015scalable} dataset which contains low recognizability examples labeled ``junk'' and ``distractors'' in its gallery set.
And likewise for partially perturbed Deepfashion~\cite{liuLQWTcvpr16DeepFashion} In-Shop dataset (most image retrieval datasets do not contain natural quality corruption, so we manually perturb the recognizability, similar to Fig. \ref{fig_arti_noise}). 
In Fig.~\ref{fig_reid_clustering}, we observe low recognizability miscellaneous samples can gather in one cluster similar to those of faces.
After devising the associated ERS measures, it can be observed from Fig.~\ref{fig_reid_RS_rank} and Fig.~\ref{fig_inshop_RS_rank} that consistent with our findings on the face, the ERS also correlates with the input image recognizability.

\subsection{Emergence of the UI cluster} \label{Emergence}
\begin{figure}[!t]
\centering
\includegraphics[width=\linewidth]{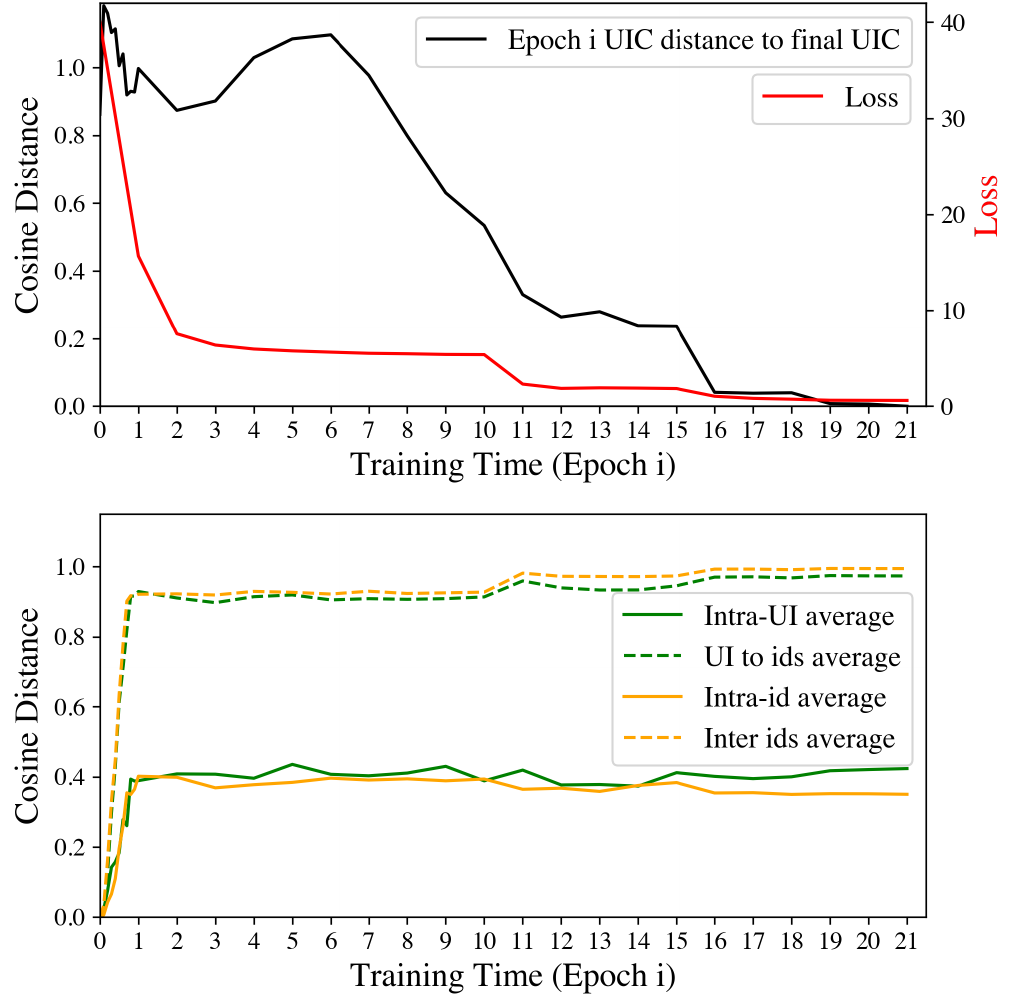}
\caption{\label{ui_during_training} Top: Trajectory of the UI centroid during training w.r.t. the final model UIC. Bottom: Distances comparison between UI images, recognizable faces from same identity and recognizable faces from different identities.}
\end{figure}

Inspired by recent works in out-of-distribution detection~\cite{huang2020feature},
we provide an exploratory study on the evolution of UI image embedding during training. We analyze the pair-wise cosine distance of embedding vectors within and between three sets of images: 
(a) UI images from known UI cluster, 
(b) recognizable faces from the same identities,
(c) recognizable faces from different identities. 
During training, we compute UIC on set (a) after each epoch.
We measure the distance of these intermediate UIC to the UIC obtained after the final epoch.
We average results from 10 independent training runs and show them in Fig.~\ref{ui_during_training}. We can see that as the training number of epochs increases: 
(1) The UI cluster emerges early on during training. Its centroid shifts significantly during model training.
(2) The distance between UI images and recognizable ids stays high, similar to that among faces from the different ids (we call these the High Distance Groups). 
(3) Distances among UI images remain at low values, similar to that of faces from the same id (Low Distance Groups)
(4) Between the High Distance Groups and the Low Distance Groups, a clear gap could be observed.

\section{Discussion}

Face recognition is subject to a vast array of issues that can affect the quality of its result. While we acknowledge the presence and importance of statistical and algorithmic biases, in this paper we focus on additional issues that exist even before any dataset is collected, and affects systems that are not trained on data. Simply put, some images contain sufficient information to ascertain that there {\em is} a
face, but insufficient to determine {\em whose} face it is. This gap is captured by the concept of ``recognizability'', which is affected by physical properties of the subject (sub-surface scattering and pigmentation, hairdo, accessories, makeup), but also extrinsic properties of the scene such as the nature and quality of the illuminant, physical properties of the sensor, calibration and pre-processing algorithms performed by the camera software, imaging conditions such as large aperture/high capture time resulting in motion blur, finite depth of field and resulting optical blur, just to mention a few. 

We also acknowledge that the principled solution to both cascading failures of the detection, as well as recognizability of the detected face, is to properly marginalize the corresponding latent variables. Since that is highly impractical, we settle to the intermediate inference of recognizability, through the proposal of an admittedly ad-hoc measure, suggested by the manifest clustering of unrecognizable identities. Because of the issues mentioned above, and also verified empirically, the UI is a highly heterogeneous cluster. It includes images subject to wildly varying nuisance variability, unlike other domains such as large-scale image classification, where optical-blurred images form a cluster separate from motion-blurred images, also separate from low-resolution images. 
We also observe that images of different populations are present in the UI.

With all due caveats in mind, we observe that explicitly accounting for recognizability through the admittedly unprincipled method we have proposed, we still achieve significant error reduction in face recognition on standard public benchmarks, and effectively allow a system to operate in an open-set setting without the complications of full-fledged open universe training.

\clearpage

\begin{appendices}
\section{Bias Analysis and Discussion}
In paper Sec. 1.3, we had a discussion about potential implications of biases, here we conduct preliminary bias analysis and show results.

\noindent\textbf{Test dataset and benchmark}
We use Morph~\cite{ricanek2006morph} dataset as our test data for the analysis, basic statistics of the dataset are shown in Table~\ref{morph_stats}.

Within each gender and ethnicity subgroup, we sample genuine and imposter 1v1 pairs to establish face verification benchmark protocol with gender and ethnicity breakdown.

\begin{table*}
\centering
\begin{tabular}{c|c|c|c|c|c|c|c|c|c} 
\hline
\multicolumn{2}{c|}{\multirow{2}{*}{}} & \multicolumn{7}{c|}{Ethnicity}                                   & \multirow{2}{*}{Total}  \\ 
\cline{3-9}
\multicolumn{2}{c|}{}                  & African & Asian & European & Hispanic & Indian & Other & Unknown &                         \\ 
\hline
\multirow{2}{*}{Gender} & Female       & 24898   & 536   & 109132   & 1880     & 66     & 82    & 10      & 136604                  \\ 
\cline{2-10}
                        & Male         & 155783  & 1150  & 99093    & 8908     & 322    & 93    & 102     & 265451                  \\ 
\hline
\multicolumn{2}{c|}{Total}             & 180681  & 1686  & 208225   & 10788    & 388    & 175   & 112     & 402055                  \\
\hline
\end{tabular}
\caption{\label{morph_stats} Morph~\cite{ricanek2006morph} dataset image statistics breakdown by Gender / Ancestry .}
\end{table*}

\noindent\textbf{Embedding models}
We test on two face embedding models previously used in benchmark experiments, one being our reference \emph{CosFace} embedding model (ResNet101 + CosFace Loss + DeepGlintFace), the other being the \emph{SC-Arc} model (ResNet101 + Sub-center Arcface Loss + DeepGlintFace) which has the best performance on IJB-C. 

\begin{figure}[b!]
\centering
\includegraphics[width=\linewidth]{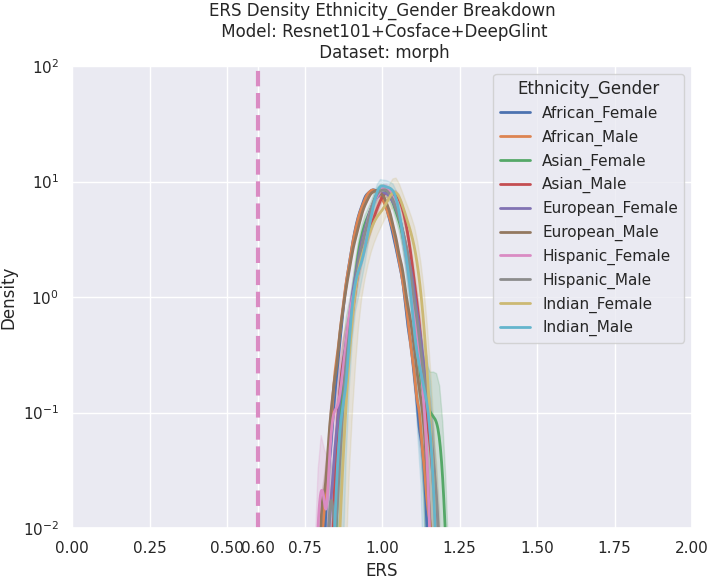}
\caption{\label{morph_density_model1} Reference CosFace model (ResNet101 + CosFace Loss + DeepGlintFace) breakdown of ERS on Morph dataset. Vertical dash indicates our threshold at 0.6.}
\end{figure}

\begin{figure}[b!]
\centering
\includegraphics[width=\linewidth]{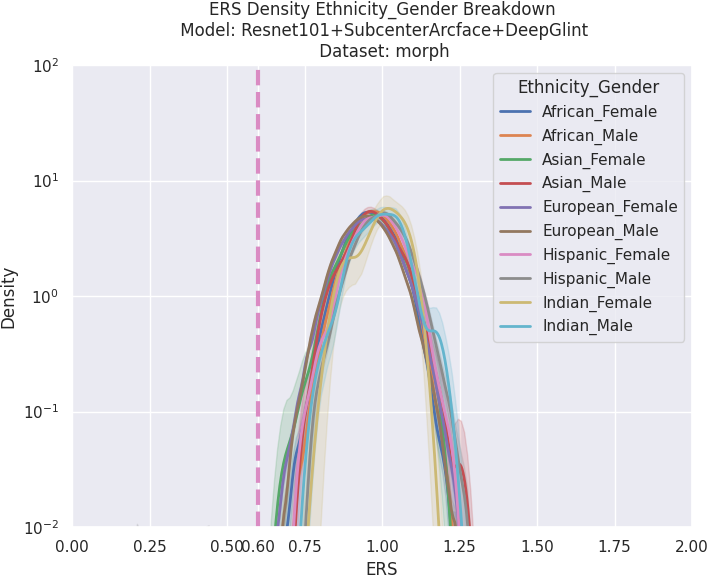}
\caption{\label{morph_density_model2} Reference SC-Arc model (ResNet101 + Sub-center Arcface Loss + DeepGlintFace) breakdown of ERS on Morph dataset. Vertical dash indicates our threshold at 0.6.}
\end{figure}

\noindent\textbf{Recognizability prediction distribution}
We plot the recognizability prediction of $e = 1 - \langle\mathbf{f}_{UI}, \mathbf{f}_i\rangle$. Comparing to ERS defined in Eq.1, the cap at 1 operation is removed for the convenience of  observing the raw distribution.
We show $e$ density distribution on Morph dataset for CosFace model in Fig.~\ref{morph_density_model1}, and for SC-Arc model in Fig.~\ref{morph_density_model2}.

\noindent\textbf{Verification benchmark}
We also test and show face verification benchmark results with gender and ethnicity breakdown on Morph dataset for CosFace model in Table.~\ref{morph_1v1_model1}, and for SC-Arc model in Table.~\ref{morph_1v1_model2}.

\begin{table}[t!]
\centering
\begin{adjustbox}{max width=\linewidth}
\begin{tabular}{c|c|c} 
\hline
FAR              & FRR@FAR=1e-2 & FRR@FAR=1e-3  \\ 
\hline
Overall          & 0.00033          & 0.00044           \\ 
\hline
African\_Female  & 0.00021          & 0.00091           \\ 
\hline
African\_Male    & 0.00015          & 0.00022           \\ 
\hline
Asian\_Female    & 0.00218          & 0.00437           \\ 
\hline
Asian\_Male      & 0.00000          & 0.00000           \\ 
\hline
European\_Female & 0.00046          & 0.00058           \\ 
\hline
European\_Male   & 0.00028          & 0.00030           \\ 
\hline
Hispanic\_Female & 0.00061          & 0.00061           \\ 
\hline
Hispanic\_Male   & 0.00020          & 0.00035           \\
\hline
\end{tabular}
\end{adjustbox}
\caption{\label{morph_1v1_model1} Reference CosFace model (ResNet101 + CosFace Loss + DeepGlintFace) breakdown 1v1 face verification benchmark on Morph dataset. Average is indicated by ``overall''. (``indian'' and ``other'' not tested due to insufficient number of images.)}
\end{table}

\begin{table}[t!]
\centering
\begin{adjustbox}{max width=\linewidth}
\begin{tabular}{c|c|c} 
\hline
FAR              & FRR@FAR=1e-2 & FRR@FAR=1e-3  \\ 
\hline
Overall          & 0.00037          & 0.00042           \\ 
\hline
African\_Female  & 0.00019          & 0.00023           \\ 
\hline
African\_Male    & 0.00018          & 0.00019           \\ 
\hline
Asian\_Female    & 0.00218          & 0.00218           \\ 
\hline
Asian\_Male      & 0.00000          & 0.00000           \\ 
\hline
European\_Female & 0.00048          & 0.00052           \\ 
\hline
European\_Male   & 0.00030          & 0.00030           \\ 
\hline
Hispanic\_Female & 0.00061          & 0.00061           \\ 
\hline
Hispanic\_Male   & 0.00025          & 0.00025           \\
\hline
\end{tabular}
\end{adjustbox}
\caption{\label{morph_1v1_model2} Reference SC-Arc model (ResNet101 + Sub-center Arcface Loss + DeepGlintFace) breakdown 1v1 face verification benchmark on Morph dataset. Average is indicated by ``overall''. (``indian'' and ``other'' not tested due to insufficient number of images.)}
\end{table}

Conclusions from the preliminary results:

(1) The different distributions may be attributed to the particularity of the model evaluated and random fluctuations in the data, rather than systematic characteristics of the design. This is evidenced by that across two models, we do not observe one group (defined by Morph dataset ethnicity or gender labels) to consistently have higher or lower ERS in general.

(2) Albeit there could be biases in these models according to verification results, 
the differences among groups only matter little in the application of the ERS we proposed. 
The proposed ERS yields high prediction ($>$0.8) for most images, regardless of the labeled gender and ethnicity. The chosen threshold 0.6 almost never generates false positive prediction for recognizability across all these groups. 

(3) The raw ERS distributions do not reveal a strong correlation with the model’s face verification performance in each subgroup. This is evidenced by, for example, ``Asian\_Female'' group has the highest error rate in both models, but its raw ERS distributions appear to be average among the curves.

It is worth noting that although the breakdown face verification results imply some biases of the face embedding models used, our investigation is far from thorough, given fairness is not the main focus of this work.
Comprehensive analysis is needed to draw more solid conclusions, for which we refer readers to the dedicated literature on face representation learning fairness analysis.


\section{Additional Experimental Results}
\noindent\textbf{Qualitative Results of ERS} 

Click \href{https://youtu.be/MNOsKIAABwk}{here} to watch a demo video on Youtube with qualitative visualization of the ERS.

\noindent\textbf{Benchmark Results on YoutubeFaces} 

We show results on verification benchmark from YoutubeFaces in Table~\ref{table_YTF}. 
Our method slightly improves the baseline: most of the faces are predicted as high recognizability (99\% of the frames have ERSs above 0.8), thus ERS aggregation is similar to average pooling. 

\begin{table}[!t]
\centering
\begin{adjustbox}{max width=0.8\linewidth}
\begin{tabular}{c|c|c}
\hline
Method        & Baseline (AvePool) & ERS   \\ \hline
Accuracy (\%) & 96.62              & 96.64 \\ \hline
\end{tabular}
\end{adjustbox}
\caption{\label{table_YTF} Templated-based face verification test on YoutubeFaces benchmark.}
\vspace{-0.5cm}
\end{table}

\noindent\textbf{Application in Person Re-Identification}

As a quick experiment to test if our method has the potential to generalize to tasks beyond face recognition, we apply our method on Market1501~\cite{zheng2015scalable} and show results in Table~\ref{table_reid}.
It can be seen that our method can improve both mAP and rank-K accuracy.


\section{More Implementation Details} \label{More_implementation_details}
\noindent\textbf{Face Clustering}

We use HAC algorithm~\cite{day1984efficient} to cluster normalized embedding features.
For reference, when running face clustering on WIDERFace validation split with features extracted with Arcface model ``MS1MV2: MS1M-ArcFace'' from \href{https://github.com/deepinsight/insightface/tree/master/model_zoo}{Insightface model zoo}.
We find distance threshold 1.0 and single linkage suitable using Scipy function  \href{https://docs.scipy.org/doc/scipy/reference/generated/scipy.cluster.hierarchy.fcluster.html}{\emph{fcluster}}. 

The optimal clustering parameters may vary from one embedding model to another, but we empirically find that the algorithm is not very sensitive to the distance threshold in order to obtain a set of unrecognizable images.
Grid searching with distance step 0.1 usually gives satisfying results, and one set of parameters can generalize to multiple models.

\begin{figure}[!t]
\centering
\includegraphics[width=\linewidth]{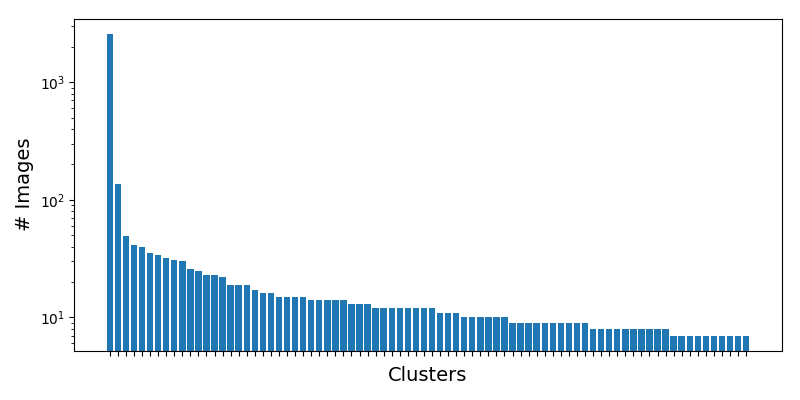}
\caption{\label{fig: cluster_hist} Clustering results on WIDER Face~\cite{yang2016wider} dataset. Cluster sizes are shown in descending order, y-axis in log-scale. The first clusters is larger than the other clusters by two or more orders of magnitude, and are full of miscellaneous low-recognizability images.}
\end{figure}

We empirically find it easy to select the UI cluster by size: The number of images in the UI cluster is larger than most of the other clusters by more than two orders of magnitude. As an example, cluster distribution on WIDERFace is shown in Fig.~\ref{fig: cluster_hist}. 

\noindent\textbf{Selection of ERS threshold}

We use images from the TinyFace~\cite{wang2014low} dataset to generate a 1v1 face verification protocol, test on it using ERS as the recognizability measure to select the best threshold (Table.~\ref{Tinyface_1v1}).
We can see $\gamma=0.60$ yields best overall performances at multiple FARs.

\begin{table}
\centering
\begin{tabular}{c|c|c|c} 
\hline
\multirow{2}{*}{}    & \multicolumn{3}{c}{FRR@FAR=}                     \\ 
\cline{2-4}
                     & 1e-2            & 1e-3            & 1e-4             \\ 
\hline
Baseline             & 0.6820          & 0.9788          & 0.9960           \\ 
\hline
ERS~($\gamma=0.50$)~ & \textbf{0.4718} & 0.5853          & 0.7133           \\ 
\hline
ERS~($\gamma=0.60$)~ & \textbf{0.4763} & \textbf{0.5772} & \textbf{0.7019}  \\ 
\hline
ERS~($\gamma=0.70$)~ & 0.4896          & \textbf{0.5768} & \textbf{0.6948}  \\ 
\hline
ERS~($\gamma=0.80$)~ & 0.5178          & 0.5871          & 0.6820           \\
\hline
\end{tabular}
\caption{\label{Tinyface_1v1} Tinyface face verification benchmark for ERS threshold selection, best two results each column are in bold. $\gamma=0.60$ yields the best overall performances at multiple FARs.}
\end{table}

\begin{table}[!t]
\centering
\begin{adjustbox}{max width=\linewidth}
\begin{tabular}{c|c|c|c|c}
\hline
\multirow{2}{*}{Setting} & \multirow{2}{*}{mAP} & \multicolumn{3}{c}{Rank-N Acc} \\ \cline{3-5} 
                         &                      & 1         & 5        & 10       \\ \hline
Baseline~\cite{zheng2019joint}                & 0.7204               & 0.8786    & 0.9525   & 0.9697   \\
ERS ($\gamma=0.70$)                 & 0.7314               & 0.8884    & 0.9569   & 0.9734   \\ \hline
\end{tabular}
\end{adjustbox}
\caption{\label{table_reid} Application on Market1501 for person re-id.}
\end{table}

\noindent\textbf{Implementation details of FaceQnet and SER-FIQ}

We followed the best practice possible to ensure the comparison with FaceQnet~\cite{hernandez2019faceqnet} and SER-FIQ~\cite{terhorst2020ser} to be fair. 
We used the \href{https://github.com/pterhoer/FaceImageQuality}{original implementation} from SER-FIQ to get prediction scores, during which we also adapt their preferred face detector and face embedding model from \href{https://github.com/deepinsight/insightface}{Insightface}.
To keep the prediction scores from SER-FIQ and face embedding model consistent, we used the same face embedding model from Insightface throughout the IJB-C Covariate Test benchmark to obtain results in the paper Table 1.
This is also a demonstration of our method working effectively on an arbitrary face embedding model not trained by us.
For FaceQnet, we also used the original implementation (https://github.com/uam-biometrics/FaceQnet) to get prediction scores, during which the preferred face detector was adapted.

\end{appendices}

{\small
\bibliographystyle{ieee_fullname}
\bibliography{arxiv}
}

\end{document}